\documentclass{article}
\usepackage{graphicx}
\usepackage[margin=1in]{geometry}
\usepackage{hyphenat}
\usepackage[utf8]{inputenc}
\usepackage[T1]{fontenc} 
\usepackage{hyperref}
\usepackage{url}
\usepackage{booktabs}
\usepackage{amsfonts}
\usepackage{nicefrac}
\usepackage{microtype}
\usepackage{lipsum}
\usepackage{fancyhdr}
\usepackage{float}
\usepackage{amsmath}
\usepackage{enumitem}
\usepackage{multirow}
\usepackage{amssymb}

\title{\textbf{MolGraphBench: A Benchmark of GNN Architectures for Molecular Regression Tasks}}

\author{ \textbf{Rajan\textsuperscript{1}} \and \textbf{Ishaan Gupta\textsuperscript{2}} }

\date{ }

\begin{document}

\maketitle

\begin{abstract} 
Molecules are often represented as SMILES strings, which can be readily converted to hand-crafted descriptors or fingerprints (FP) for molecular property prediction. Research has demonstrated that SMILES can be converted to molecular graphs $G = (V, E)$, with atoms as nodes $(V)$ and bonds as edges $(E)$. These molecular graphs can subsequently be used to train graph neural networks (GNN) models. Despite the recent surge in application of GNN (existing and novel architectures) for molecular property prediction, a rigorous benchmark is still lacking. We propose MolGraphBench (\url{https://github.com/rajanbit/MolGraphBench}), a comprehensive benchmark of four commonly used GNN models for molecular property prediction. Benchmarking results demonstrate graph convolutional network (GCN) and graph isomorphism networks (GIN) as the optimal GNN architectures for molecular graph regression tasks, based on absolute performance, training efficiency, transfer learning and prediction quality. The study also indicates the non-complementary nature of molecular fingerprints in the fusion (GNN-FP) framework. Furthermore, our GNN models achieved performance superior or comparable performance to current state-of-the-art GNN baselines across three datasets (GCN with RMSE of $0.518$ on B3DB, GIN-FP with RMSE of $1.022$ on FreeSolv and GIN with MAE of $63.783$ on RT datasets). Findings from this study indicate that type of GNN-layer, should be treated as a tunable hyperparameter rather than a fixed design choice to achieve superior performance.

\centering
\includegraphics[width=0.9\textwidth]{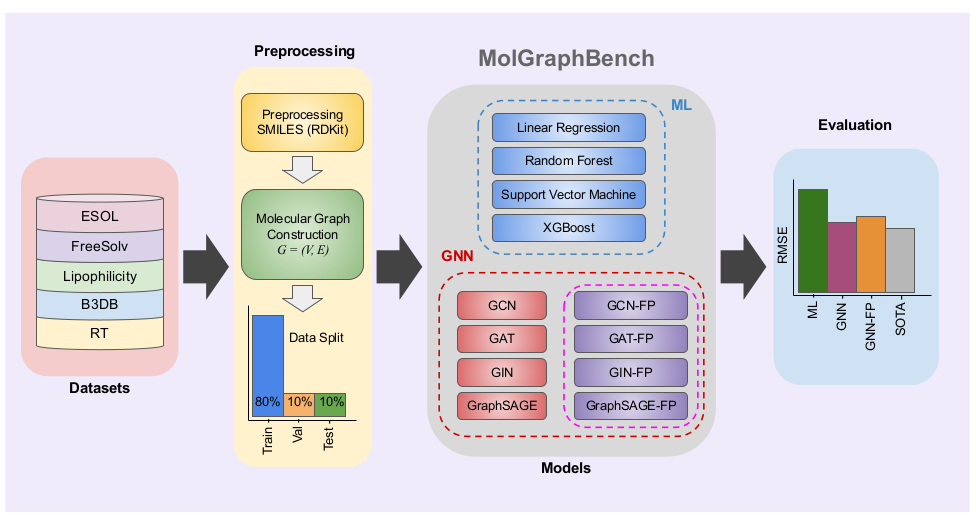}
\par\medskip
\end{abstract}


\footnotetext[1]{School of Interdisciplinary Research (SIRe), IIT Delhi, New Delhi-110016, India}
\footnotetext[2]{Department of Biochemical Engineering and Biotechnology (DBEB), IIT Delhi, New Delhi-110016, India}

\section{Introduction}
\label{sec:Introduction}

Molecular property prediction is a critical task in drug discovery, computational and analytical chemistry \cite{Zagidullin2021, Lutchyn2025, Liu2025}. While many such properties are numerical (e.g. solubility and binding affinity), some of the properties can be expressed as discrete categories (e.g. toxicity and permeability) \cite{Wang2023}. Traditionally these properties are predicted using manually crafted fixed-length chemical descriptors and molecular fingerprints. Machine learning (ML) and deep learning (DL) models like linear regression (LR),  random forest (RF), support vector machine (SVM), XGBoost (XGB), multilayer perceptron (MLP) and artificial neural network (ANN) can be trained using these descriptors or fingerprints for molecular property prediction \cite{Liu2025, Wang2023}. These ML/DL approaches often rely on expert-driven feature engineering or high dimensional  (1024 or 2048 bits) sparse molecular fingerprints, thus requiring large data to avoid overfitting and offering limited explainability \cite{Yang2019}. Moreover, these models are task dependent and require complete model training for predicting new-property \cite{Wozniak2025}.

\vspace{1em}

Recently, geometric deep learning (GDL) models have gained significant adoption for molecular property prediction. Since graph-based representations are more obvious to molecules, GDL models like graph neural networks (GNN) outperform conventional ML/DL approaches that depend on molecular descriptors and fingerprints \cite{Lutchyn2025, Jiang2024, Sun2025, Li2025, Kumar2025, Ahmad2023, Zhang2024, Buterez2024, Nguyen2026, Yang2021}. GNN models are typically trained on molecular graphs, where nodes represent atoms and edges correspond to bonds. These models learn unique molecular embeddings through a message-passing mechanism, where, node (atom) embeddings are updated using the aggregated neighbors (adjacent atoms) embeddings, thereby allowing these models to learn both local and global context \cite{Sun2025}. In contrast to ML/DL models trained on heuristics driven molecular fingerprints, GNN models are learnable, therefore, higher performance can be achieved with the development of better architectures \cite{Li2025, Altae-Tran2017}. There are many GNN architectures like classical GCN (Graph Convolutional Network), attention based GAT (Graph Attention Network), inductive GraphSAGE (Graph Sample and Aggregate), highly expressive GIN (Graph Isomorphism Network), general MPNN (Message-Passing Neural Network) and transformer based Graphormer (Graph Transformer) \cite{Kipf2017, Velikovi2018, Hamilton2018, Xu2019, Gilmer2017, Ying2021}. These architectures differ in their aggregation and update function during neural message-passing \cite{Gilmer2017}. GCN, GAT, GraphSAGE and GIN are the representative and widely established GNN architectures used for molecular property prediction \cite{Zhang2019, Gaudelet2021}.

\vspace{1em}

Additionally, novel GNN architectures have also been proposed by incorporating the baseline variants (GCN, GAT, GraphSAGE and GIN), either alone or in combination (Table \ref{tab:tab1}). GNN-TL, MvMRL, MolGraph-xLSTM and graphB3 incorporates GCN, whereas FP-GNN, AttentiveFP and  TChemGNN uses GAT for molecular graph convolution \cite{Lutchyn2025, Sun2025, Kumar2025, Ahmad2023, Zhang2024, Yang2021, Cai2022}. KA-GNNs (Kolmogorov–Arnold GNNs) and DGCL (dual graph NN for contrastive learning) use a combination of GAT with, either GCN or GIN \cite{Jiang2024, Li2025}. Models like D-MPNN, FH-GNN and GMC-MPNN use MPNN architecture that incorporates both nodes (atoms) and edges (bonds) features during graph convolution \cite{Liu2025, Yang2019, Nguyen2026}. Finally, there are transformer based architectures like CLAPS \cite{Wang2023}. As can be seen in Table \ref{tab:tab1} GCN and GAT as the most preferable choice for model development.

\vspace{1em}

\begin{table}[htbp]
\centering
\caption{Details of novel GNN architectures for molecular property prediction.}
\label{tab:tab1}
\vspace{0.2cm}
\begin{tabular}{l c l l l c}
\hline
\textbf{Authors} & \textbf{Year} & \textbf{Model} & \textbf{GNN Type} & \textbf{Code Availability (GitHub)} & \textbf{Ref.} \\
\hline
Yang et al. & 2019 & D-MPNN & MPNN & chemprop/chemprop & \cite{Yang2019} \\
Yang et al. & 2021 & GNN-TL & GCN & Qiong-Yang/GNN-TL & \cite{Yang2021} \\
Cai et al. & 2022 & FP-GNN & GAT & idrugLab/FP-GNN & \cite{Cai2022} \\
Ahmad et al. & 2023 & AttentiveFP & GAT & waqarahmadm019/AquaPred & \cite{Ahmad2023} \\
Wang et al. & 2023 & CLAPS & Transformer & wangjx22/CLAPS & \cite{Wang2023} \\
Jiang et al. & 2024 & DGCL & GAT \& GIN & Sysuzqs/DGCL & \cite{Jiang2024} \\
Zhang et al. & 2024 & MvMRL & GCN & jedison-github/MvMRL & \cite{Zhang2024} \\
Sun et al. & 2025 & MolGraph-xLSTM & GCN & syan1992/MolGraph-xLSTM & \cite{Sun2025} \\
Li et al. & 2025 & KA-GNNs & GCN \& GAT & LongLee220/KA-GNN & \cite{Li2025} \\
Kumar et al. & 2025 & graphB3 & GCN & dhanjal-lab/graphB3 & \cite{Kumar2025} \\
Lutchyn et al. & 2025 & TChemGNN & GAT & uitml/TChemGNN & \cite{Lutchyn2025} \\
Liu et al. & 2025 & FH-GNN & MPNN & shuoliu0-0/FH-GNN & \cite{Liu2025} \\
Nguyen et al. & 2026 & GMC-MPNN & MPNN & MathIntelligence/GMC-MPNN-BBBP & \cite{Nguyen2026} \\
\hline
\end{tabular}
\end{table}

\begin{figure}[H]
\centering
\includegraphics[width=0.95\textwidth]{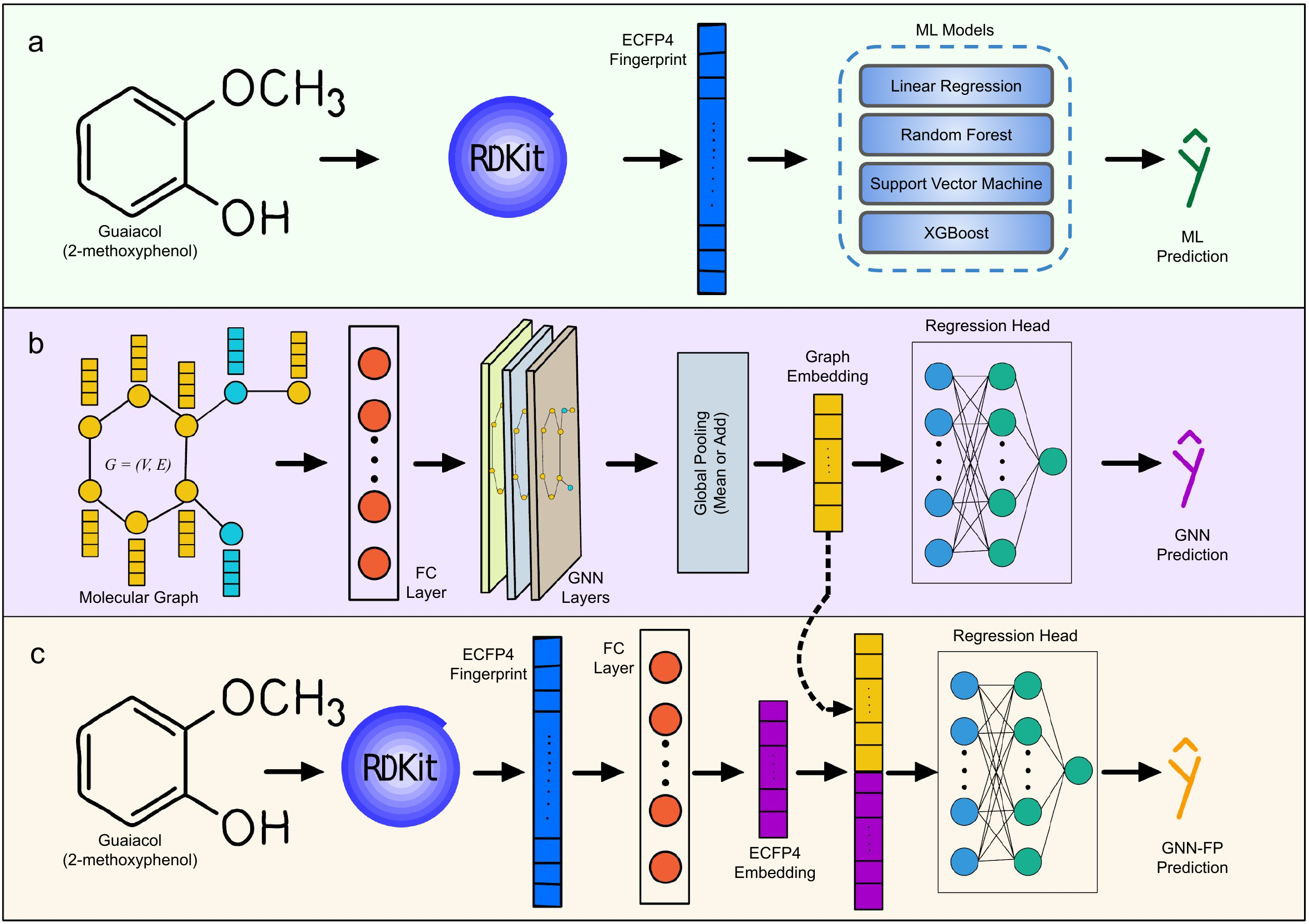}
\caption{ \textbf{Schematic representation of MolGraphBench benchmark.} (a) Description of ML models trained using ECFP4 fingerprints. (b) Brief summary of GNN models trained using molecular graphs. (c) Fusion GNN-FP architecture trained using GNN and ECFP4 fused embeddings.
 }
\label{fig:fig1}
\end{figure}

Despite their success, research indicates that lack of relevant and impactful benchmarks can potentially hinder future developments within the field. With current benchmarks mainly focusing on predictive performance and narrow domain adaptation, the generalization stability and its real-world application remains questionable \cite{Bechler-Speicher2025, Zhang2026}. The present study proposes a comprehensive benchmark of most common GNN architectures (GCN, GAT, GraphSAGE and GIN) for molecular regression tasks (Figure \ref{fig:fig1}). Here GNN models were compared against the ML baselines and fingerprint enabled GNN variants. All the developed models were benchmarked against five regression datasets (ESOL, FreeSolv, Lipophilicity, B3DB and RT) for their absolute performance. In addition to their predictive performance; training time, prediction quality, generalizability and explainability were also evaluated. Findings from this benchmark study can help researchers make better decisions while selecting GNN architectures for molecular regression tasks.

\section{Methods}
\label{sec:Methods}

\subsection{Datasets and Preprocessing}
Physical chemistry datasets (ESOL, Lipophilicity and FreeSolv) were downloaded from the MoleculeNet database \cite{Wu2018}. The other two datasets, B3DB and RT were obtained from the published literature \cite{Yang2021, Meng2021}. For all the datasets SMILES strings were preprocessed for tautomer standardization, neutralization of charged species, and removal of counterions and salts using RDKit library \cite{Bento2020}. The processed datasets were randomly divided into training ($80\%$), validation ($10\%$) and hold-out test ($10\%$) \cite{Sun2025}. The training and validation sets were used for hyperparameter optimization and model training. The hold-out test set was used to evaluate the trained models with optimal hyperparameters.

\subsection{Baseline ML Models}
A robust baseline was constructed using four machine learning models – LR, SVM, RF, and XGB regressors. The baseline ML models were trained using the 1024-bit ECFP4 fingerprints calculated using RDKit library as done by Wu et al., 2018 \cite{Wu2018, Bento2020}. For each model, hyperparameter optimization was performed independently on all datasets and optimal-hyperparameters were selected based on the lowest root mean square error (RMSE) value.

\subsection{Graph Neural Network Models}

\subsubsection{Molecular Graph Construction}
The molecular graph has been constructed as $G = (V, E)$ be an undirected graph where nodes $V$ are atoms and edges $E$ are chemical bonds. Each node $v \in \text{V}$ is associated with a feature vector $X_v$ calculated as described by Sun et al., 2025 \cite{Sun2025}, resulting in an input features matrix $X \in \mathbb{R}^{N \times 118}$. The structure of the molecular graph $(G)$ is represented by an adjacency matrix $A \in \mathbb{R}^{N \times N}$, where $N$ is the number of atoms present in the molecule.

\subsubsection{Model Architecture}
The GNN model consists of a fully-connected (FC) projection layer that maps input feature matrix $X \in \mathbb{R}^{N \times 118}$ into hidden dimension $X_h \in \mathbb{R}^{N \times H}$. 

\begin{equation}
    X_h = ReLU(X W_{(FC)})
\end{equation}

where $W_{(FC)} \in \mathbb{R}^{118 \times H}$

\vspace{1em}

This is followed by graph convolution layers (2-3 layers) containing GraphNorm and residual connection to prevent vanishing gradients. ReLU activation function was used to add non-linearity between the layers. The convolution layer can be a GCN, GAT, GIN or GraphSAGE.

\vspace{1em}

For layers $k = 1, \dots, L$, each node $v \in \text{V}$ updates its representation $h_v^{(k)}$:

\begin{equation}
    m_v^{(k+1)} = \text{AGGREGATE}^{(k+1)} \left( \{ h_u^{(k)} : u \in \mathcal{N}(v) \cup \{v\} \} \right)
\end{equation}
\begin{equation}
    h_v^{(\hat{k})} = \text{UPDATE}^{(k+1)} \left( h_v^{(k)}, m_v^{(k+1)} \right)
\end{equation}

\begin{equation}
    \tilde{h}_v^{(k)} = \text{ReLU}\left( \text{GraphNorm}\left( h_v^{(\hat{k})} \right) \right)
\end{equation}
\begin{equation}
    h_v^{(k)} = \tilde{h}_v^{(k)} + h_v^{(k-1)}
\end{equation}

 Following graph convolution, node embeddings were aggregated to generate graph-level embedding $(G_{embed})$ using global mean pooling (or global add pooling in case of GIN layers \cite{Xu2019}, which is subsequently processed by a two-layer MLP regression head with single output neuron for molecular property prediction. 

\begin{equation}
    G_{embed} = \text{POOL} \left( \{ h_v^{(L)} : v \in \mathcal{V} \} \right)
\end{equation}
where $\text{POOL}$ is $\sum_{v \in \mathcal{V}}$ for GIN, and $\frac{1}{|\mathcal{V}|} \sum_{v \in \mathcal{V}}$ otherwise.

\begin{equation}
    \hat{y} = MLP(G_{embed})
\end{equation}
 
 The GNN-FP variants have an additional projection layer that maps 1024-bits ECFP4 fingerprints into hidden dimension $(F_{embed})$, which is concatenated with graph-level embedding $(G_{embed})$ before being fed to the two-layer MLP regression head for molecular property prediction. The brief description of GNN and GNN-FP models is shown in Figure \ref{fig:fig1}. 

\subsubsection{Hyperparameter Optimization and Model Training}
Hyperparameter optimization for all GNN and GNN-FP models were independently performed for all datasets. Optimal hyperparameters were selected based on the lowest RMSE value on the validation set. Each model was finally trained for 1000 epochs using the Adam optimizer with mean squared error (MSE) as the loss function and optimized hyperparameter. For hyperparameter optimization and model training an early stopping criterion (with patience of 20 epochs) based on validation loss was used. 

\subsection{Prediction Error Analysis} 
We have also quantified the percentage of molecules which have high prediction error i.e. $\left| y_{\text{test}} - y_{\text{pred}} \right| > \mathrm{SOTA}_{\text{threshold}}$ (absolute prediction error more than SOTA i.e. state-of-the-art models) for each model (ML and GNN). Additionally, we studied the molecules with high prediction error and compared them against the molecules with low prediction error i.e. $\left| y_{\text{test}} - y_{\text{pred}} \right| \leq \mathrm{SOTA}_{\text{threshold}}$ (absolute prediction error less than or equal to SOTA) for all the top performing ML and GNN models. The molecules were compared for six molecular features – molecular weight, number of atoms, number of bonds, ring count, H-donors and H-acceptors.

\subsection{Transfer Learning With GNN}
We further investigate the potential of GNN models for transfer learning, where all the GNN models were initially trained on METLIN datasets (training approach same as mentioned in section 2.3) \cite{Domingo-Almenara2019}. The trained GNN models were then fine-tuned on the RT dataset (training and validation set) and finally evaluated on the RT hold-out test set. Three levels of fine-tuning were performed which are as follows: fine-tuning regression (two-layer MLP) head, fine-tuning the FC projection layer in addition to regression head and, finally complete model fine-tuning. All three fine-tuning experiments were performed by training models for 100 epochs with early stopping criteria (patience = 10 epochs) based on validation loss using Adam optimizer with MSE loss.

\subsection{Explaining GNN Predictions}
The prediction for GNN models can also be explained using libraries such as GNNExplainer \cite{Ying2019}. We applied GNNExplainer to seven case-study molecules. These seven molecules were selected because they were common within the hold-out test set of atleast two datasets. This was done to investigate task-specific molecular features (e.g. distinguishing atom and bond pairs within the same molecule which are most important for hydration free energy i.e. FreeSolv or log solubility i.e. ESOL). We performed predictions and generated feature importance masks for these seven molecules across distinct targets using task specific best models (e.g. GraphSAGE model trained on ESOL data). The importance mask was overlaid to the molecule structure for final visualization.

\subsection{Evaluation Metrics}
All models (ML baselines, GNN, and GNN-FP) were assessed using RMSE as the evaluation metric. To ensure the robustness and reproducibility all experiments were run in triplicate (n=3) and we report the RMSE using the mean and standard deviation across these independent runs. For the RT dataset the SOTA performance was reported in mean absolute error (MAE), therefore in addition to RMSE, we have also calculated MAE for all the analysis done on RT datasets to draw a fair comparison with SOTA, wherever necessary.

\begin{equation}
\text{RMSE} = \sqrt{\frac{1}{n} \sum_{i=1}^{n} (y_i - \hat{y}_i)^2}
\end{equation}

\begin{equation}
\text{MAE} = \frac{1}{n} \sum_{i=1}^{n} \left| y_i - \hat{y}_i \right|
\end{equation}

\vspace{1em}

The codes were written in Python, with data analysis and experimentation conducted within Jupyter Notebooks. The models were trained on a high-performance computing (HPC) server equipped with NVIDIA RTX A6000 GPU (48 GB vRAM), a 64-core CPU, and 315 GB of RAM. Throughout the experimental phase, GPU utilization peaked at approximately $80\%$. ML baseline models were implemented using scikit-learn and xgboost library. GNN and GNN-FP architectures were implemented using PyTorch and PyTorch-geometric libraries.

\vspace{1em}

The ESOL and Lipophilicity datasets can be downloaded from MoleculeNet (\url{https://moleculenet.org/datasets-1}). The RT and B3DB datasets are available from their respective GitHub repositories (\url{https://github.com/Qiong-Yang/GNN-TL} and \url{https://github.com/theochem/B3DB}). The processed datasets and codes used in this study are available on the official GitHub repository (\url{https://github.com/rajanbit/MolGraphBench}).

\section{Results and Discussion}
\label{sec:Results_and_Discussion}

\subsection{Datasets Summary and SOTA Benchmarks}
\begin{figure}[H]
\centering
\includegraphics[width=0.95\textwidth]{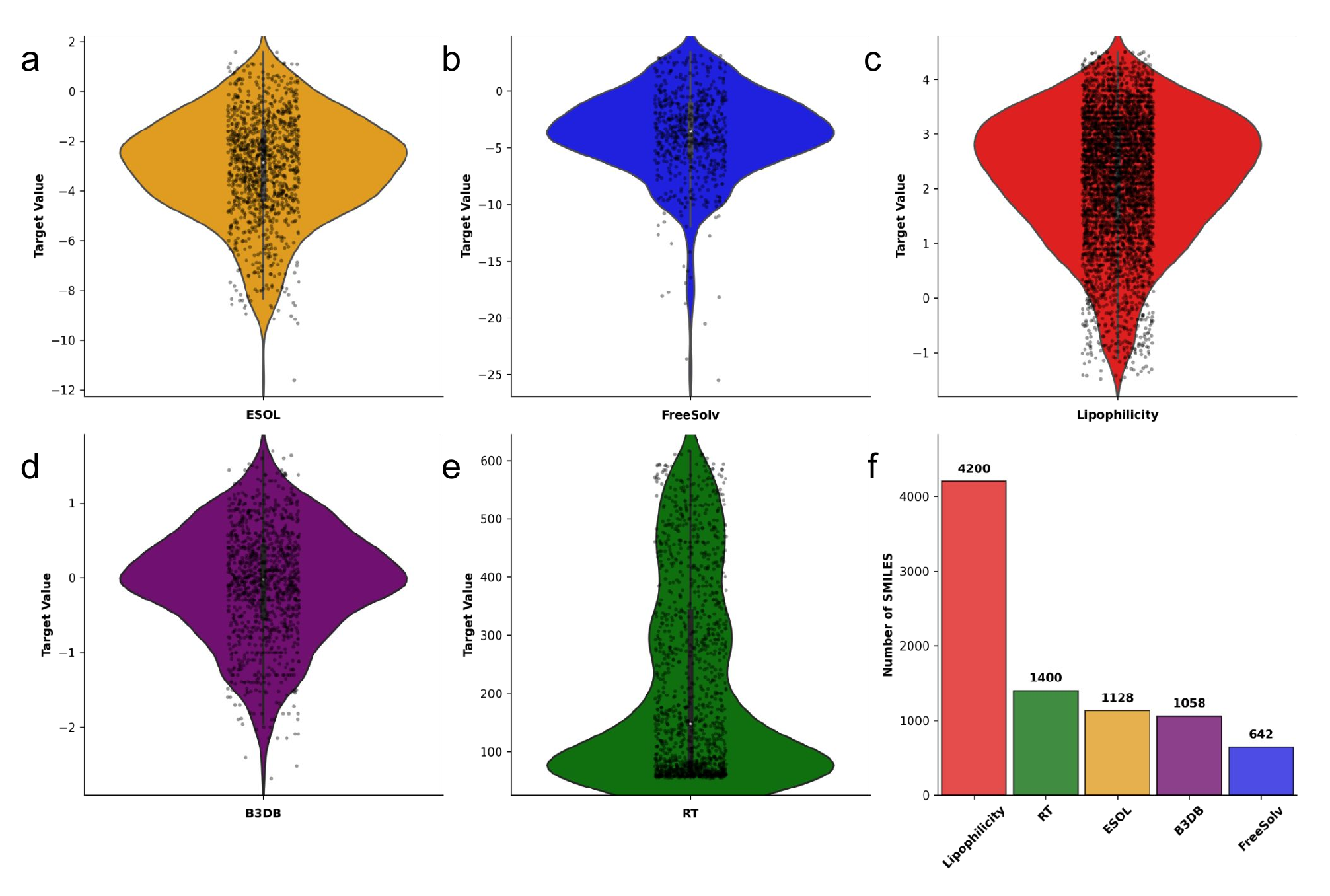}
\caption{\textbf{Data summary plot.} (a-e) Violin plots showing target distribution across datasets. (f) SMILES count for each dataset}
\label{fig:fig2}
\end{figure}

We incorporated three datasets from the physical chemistry domain, and one dataset from each, biological and analytical domain. This was done to investigate the domain level difference in performance of different ML and GNN models. The ESOL dataset consists of experimental aqueous solubility values for 1028 molecules. The FreeSolv dataset provides hydration free energy of 642 molecules. Lipophilicity dataset contains experimentally measured logD values for 4200 molecules. All these three datasets (ESOL, FreeSolv and Lipophilicity) belong to the physical chemistry domain. The B3DB dataset (biological) contains a curated list of 1058 blood-brain barrier permeable molecules with their logBB values, and RT dataset (analytical) contains retention-time measurements obtained from hydrophilic interaction liquid chromatography (HILIC) for 1400 molecules. The distribution of molecular properties for each dataset is shown in Figure \ref{fig:fig2}(a-e) as a violin plot, with barplot (Figure \ref{fig:fig2}f) showing the size of datasets in decreasing order.

The performance of all the models developed in this study has been compared against SOTA models. The SOTA models were either published in the original study (GNN-TL for RT dataset) or have been proposed recently (MolGraph-xLSTM\cite{Sun2025} for ESOL, FreeSOLV, Lipophilicity datasets and GMC-MPNN\cite{Nguyen2026} for B3DB dataset). The SOTA performance was reported in RMSE (except for GNN-TL\cite{Yang2021} which was reported in MAE).

\subsection{Performance of GNN Models for Molecular Property Prediction}
The performance of GNN (GCN, GAT, GraphSAGE and GIN) and GNN-FP (GCN-FP, GAT-FP, GraphSAGE-FP and GIN-FP) model variants were evaluated against the ML (LR, SVM, RF and XGB) baselines. The performance of each model was reported in RMSE values (Table \ref{tab:tab2}), where lower values indicate higher predictive performance and vice-versa. Top baseline performance was achieved using SVM model (or RF model on RT dataset), whereas LR was the worst performing baseline ML model. The baseline models were trained using the 1024-bits ECFP4 fingerprints. On the B3DB dataset the SVM model achieved the lowest RMSE (0.471) value in comparison to GNN, GNN-FP model variants as well as the SOTA model. The performance for top performing GNN and GNN-FP variants were comparable to the SVM model, with RMSE values of 0.518 and 0.508, for GCN and GCN-FP, respectively. The GNN-FP (GIN-FP with RMSE of 1.022) model was the top performing model for the FreeSolv dataset, and its performance was equivalent to that of the SOTA model. FreeSolv is the smallest dataset used in our study, and an improvement in performance was seen from GNN to GNN-FP variants (except for GCN and GCN-FP). This suggest complementary nature of ECFP4 fingerprints in GNN-FP variants in case of FreeSolv data. The GNN (GraphSAGE) models were the top performing models with RMSE value of 0.762, 0.629 and 89.417 for ESOL, Lipophilicity and RT datasets, respectively. On RT dataset GAT-FP has the lowest RMSE of 88.417 (comparable with GNN i.e. GraphSAGE with RMSE of 89.417), but, with a higher standard deviation (1.662), therefore GraphSAGE was considered the top performer on this dataset. Additionally, to draw a better conclusion for the RT dataset we have also computed the MAE values for GNN variants. Based on the MAE value, the GIN was the top performing model with MAE of 63.783, whereas GraphSAGE was the second-best model with MAE of 68.311. Therefore, the GNN models (GIN and GraphSAGE) achieved superior performance on RT dataset, as compared to the SOTA. 

\vspace{1em}

The centered kernel alignment (CKA) scores between GNN and FP embeddings are between 0.27 to 0.59. The low CKA score suggests ECFP4 and GNN embeddings to be complementary in nature but GNN-FP results are counterintuitive \cite{Welsch2024}. The CKA score was also calculated to quantify embeddings similarity between GNN architecture. The CKA scores were high (CKA >0.7), except for GIN (vs other GNN) architecture, across all datasets \cite{Welsch2024}. The low CKA score of GIN vs other GNN comparison can be due to the difference in pooling formula (GIN has add pooling whereas other GNN models have mean pooling). 

\begin{table}[htbp]
\centering
\caption{Comparison of performance (RMSE with standard deviation) across different model architectures and datasets.}
\label{tab:tab2}
\vspace{0.2cm}
\begin{tabular}{l c c c c c}
\hline
\textbf{Models} & \textbf{B3DB} & \textbf{ESOL} & \textbf{FreeSolv} & \textbf{Lipophilicity} & \textbf{RT} \\
\hline
LR & 1.426 (0.000) & 4.380 (0.000) & 2.116 (0.000) & 1.011 (0.000) & 601.600 (0.000) \\
SVM & 0.471 (0.000) & 1.128 (0.000) & 1.512 (0.000) & 0.730 (0.000) & 120.555 (0.000) \\
RF & 0.550 (0.002) & 1.569 (0.009) & 2.572 (0.015) & 0.950 (0.001) & 103.548 (0.456) \\
XGB & 0.580 (0.001) & 1.615 (0.003) & 2.420 (0.006) & 1.057 (0.001) & 108.227 (0.387) \\
GCN & 0.518 (0.007) & 0.808 (0.030) & 1.288 (0.104) & 0.664 (0.010) & 96.170 (2.743) \\
GAT & 0.552 (0.011) & 0.808 (0.081) & 1.736 (0.061) & 0.639 (0.028) & 94.734 (4.295) \\
GIN & 0.555 (0.011) & 0.772 (0.044) & 1.461 (0.176) & 0.672 (0.020) & 92.344 (3.042) \\
GraphSAGE & 0.542 (0.011) & 0.762 (0.041) & 1.352 (0.071) & 0.629 (0.006) & 89.477 (1.067) \\
GCN-FP & 0.508 (0.007) & 1.000 (0.030) & 1.346 (0.064) & 0.794 (0.017) & 94.879 (1.925) \\
GAT-FP & 0.525 (0.014) & 0.979 (0.078) & 1.355 (0.081) & 0.682 (0.016) & 88.417 (1.662) \\
GIN-FP & 0.549 (0.020) & 0.814 (0.015) & 1.022 (0.067) & 0.699 (0.011) & 91.739 (0.908) \\
GraphSAGE-FP & 0.529 (0.011) & 0.969 (0.012) & 1.247 (0.064) & 0.724 (0.013) & 89.927 (1.361) \\
SOTA & RMSE: 0.56 & RMSE: 0.53 & RMSE: 1.02 & RMSE: 0.55 & MAE: 70.40 \\
\hline
\end{tabular}
\end{table}

\subsection{Comparison of Training Time Across Model Architectures}
\begin{figure}[H]
\centering
\includegraphics[width=0.95\textwidth]{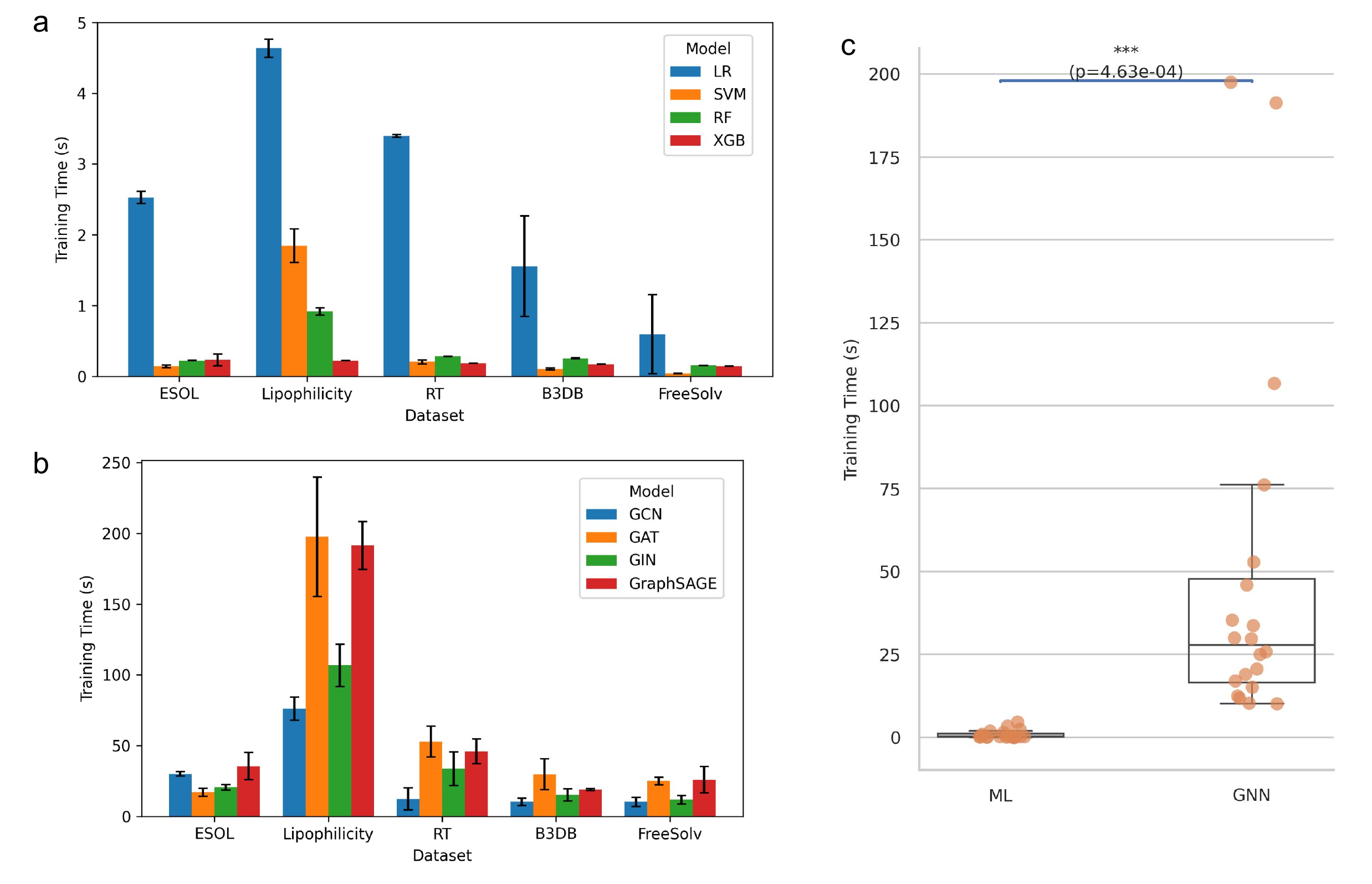}
\caption{\textbf{Benchmarking training efficiency across model architectures.} (a) Training time for baseline ML models. (b) GNN models training time. (c) Boxplot showing difference in training time for ML and GNN models. The error bar denotes standard deviation.}
\label{fig:fig3}
\end{figure}

Training time for all the models (ML and GNN) were also compared to study the training efficiency. LR was consistently the slowest trainer across all the datasets. The other three ML models (SVM, RF and XGB) exhibited comparable training time (Figure \ref{fig:fig3}a), with an exception on the lipophilicity dataset. On lipophilicity data, the scale of training time for all the ML models were significantly different, with LR and XGB as the least and most efficient, respectively. Overall, the training time for GNN models was significantly high compared to ML models (Figure \ref{fig:fig3}c). Among the GNN architectures, GCN and GIN were the fastest trainers across all the datasets (Figure \ref{fig:fig3}b). 

\subsection{Evaluating Prediction Quality of GNN and ML Models}
We further investigated the scale of error being made (i.e. absolute difference between the predicted and target values) by each model. We systematically compared the models by grouping their prediction errors into two categories (high and low errors) based on the SOTA threshold. Thus, quantifying the percentage of molecules falling into the two categories. Table \ref{tab:tab3} presents the percentage ($\%$) of high-error molecules across all ML and GNN models. A general trend can be observed from Table \ref{tab:tab1}, that the top performing models are the ones which have the lowest proportion ($\%$) of high error molecules (Table \ref{tab:tab3}). There were few exceptions to this trend – RT dataset (with RF and GraphSAGE top model based on absolute performance) and GraphSAGE model for ESOL data. On the RT dataset SVM has the lowest proportion of ($\%$) of high error molecules but RF was the top performing model based on RMSE. Similarly, for GNN variants, GIN has the lowest proportion of ($\%$) of high error molecules but GraphSAGE was the top performer on RT dataset. On ESOL data, the performance of the top performing model (GraphSAGE) has comparable ($\%$) of high error molecules with that of GIN that has lowest ($\%$) of high error molecules. Overall, this trend can be accepted with the RT dataset as an exception. Furthermore, GNN models generally have a lower percentage of high error molecules, except for the B3DB dataset (SVM with lowest $\%$ of high-error molecules). This indicates that apart from RMSE based performance metric, the prediction quality of GNN models was also better.

\begin{table}[htbp]
\centering
\caption{Percentage (standard deviation) of high error molecules across all models (ML and GNN) and datasets.}
\label{tab:tab3}
\vspace{0.2cm}
\begin{tabular}{l c c c c c}
\hline
\textbf{Models} & \textbf{B3DB} & \textbf{ESOL} & \textbf{FreeSolv} & \textbf{Lipophilicity} & \textbf{RT} \\
\hline
LR & 50.943 (0.000) & 83.186 (0.000) & 67.692 (0.000) & 54.762 (0.000) & 80.714 (0.000) \\
SVM & 18.868 (0.000) & 53.982 (0.000) & 41.538 (0.000) & 36.905 (0.000) & 48.571 (0.000) \\
RF & 30.503 (0.889) & 77.581 (1.104) & 75.897 (1.919) & 52.381 (0.514) & 60.714 (1.010) \\
XGB & 33.648 (1.177) & 77.876 (1.252) & 69.231 (0.000) & 60.159 (0.736) & 68.095 (0.891) \\
GCN & 25.472 (0.770) & 42.478 (0.723) & 32.308 (1.256) & 37.381 (0.701) & 36.905 (3.212) \\
GAT & 25.157 (1.603) & 46.018 (4.395) & 38.974 (4.038) & 38.095 (0.673) & 40.476 (0.891) \\
GIN & 27.673 (1.177) & 37.463 (0.834) & 38.462 (2.512) & 37.381 (1.010) & 32.381 (1.875) \\
GraphSAGE & 30.818 (0.445) & 40.708 (4.335) & 50.769 (2.176) & 35.714 (1.854) & 40.238 (1.468) \\
\hline
\end{tabular}
\end{table}

\begin{figure}[H]
\centering
\includegraphics[width=0.95\textwidth]{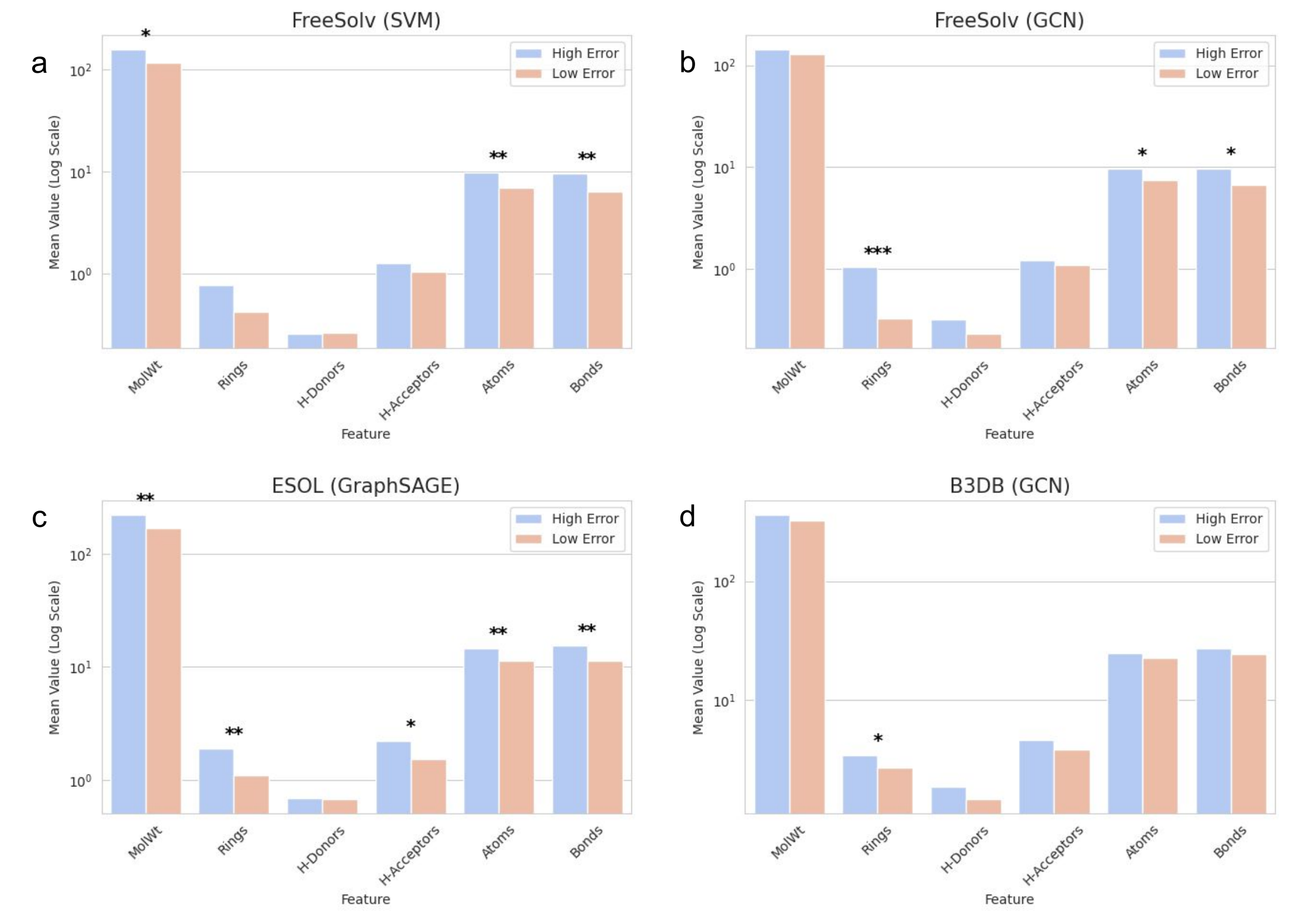}
\caption{\textbf{Difference in molecular properties across high and low error molecules.} (a) Top performing ML (SVM) model for FreeSolv dataset. (b-d) Top performing GNN (GCN and GraphSAGE) models for FreeSolv, ESOL and B3DB datasets.}
\label{fig:fig4}
\end{figure}

Moreover, we also quantified the molecular features (atomic mass, number of atoms and bonds, ring counts, H-donors and H-acceptors) for high and low error molecules. For simplicity, this was done for all the top performing ML (SVM for ESOL, FreeSolv, Lipophilicity and B3DB; and RF for RT) and GNN (GCN for B3DB and FreeSolv; and GraphSAGE for ESOL, Lipophilicity and RT) models. We observed significant differences in molecular properties for FreeSolv (SVM and GCN models), ESOL (GraphSAGE) and B3DB (GCN) datasets. In ML models only SVM classifiers for FreeSolv data (Figure \ref{fig:fig4}a) showed a significant difference in molecular property (atom and bond counts that represent the scale of the molecule). This is similar to the GCN model for FreeSolv data (Figure \ref{fig:fig4}b) where molecules with highest errors contain more atoms and bonds (larger molecules) in addition to the number of rings. For GNN models the ring counts were significantly altered between high and low error molecules for 3 out of 5 datasets (Figure \ref{fig:fig4}b, \ref{fig:fig4}c and \ref{fig:fig4}d), i.e. molecules with highest errors containing more rings. On ESOL data the GNN models (GraphSAGE) show significant alteration (Fig. \ref{fig:fig4}c) between 4 out of 5 molecular properties. The difference in molecular properties between high and low error molecules for the remaining 6 models was not significant. These results indicate that GNNs struggle for larger and complex molecules (more atoms, bonds and rings), primarily due to their reliance on molecular graphs, with diverse size (atoms and bonds) and shape (linear chain, chains, etc.). This is opposite to the ECFP4 fingerprints which have an identical dimension to represent each molecule.

\subsection{Comparison of GNN Architecture for Transfer Learning}
We have also benchmarked the transfer learning ability of all the four GNN architectures. All the GNN models were initially trained on METLIN, a retention time prediction dataset acquired using reverse-phase chromatography. It consists of 80k SMILES with their retention time. The trained model was further fine-tuned on RT dataset, at three different levels (MLP, MLP+FC and complete model). The results show the GIN model consistently achieving lowest MAE across all the three levels of fine-tuning (Table \ref{tab:tab4}), with the complete model fine-tuning achieving the lowest MAE of 52.914 (RMSE of 78.742). The second-best model was GraphSAGE for all the three levels of fine-tuning. Even with zero-shot learning (fine-tuning MLP regression head), MAE of 60.960 was achieved which is comparatively lower than the top performing GraphSAGE models (MAE of 68.311), trained and validated on RT data. The original study done by Yang et al., 2021 proposes GNN-TL, a GNN model that was trained on transfer-learning capability by training on 305k in-silico HILIC data\cite{Yang2021}. GNN-TL achieved a MAE of 38.6 on the hold-out test set for RT data after fine-tuning. In our case we used METLIN which is for reverse-phase chromatography that has a totally different distribution.

\begin{table}[htbp]
\centering
\caption{Transfer learning performance (MAE with standard deviation) of GNN models for RT dataset.}
\label{tab:tab4}
\vspace{0.2cm}
\begin{tabular}{l c c c}
\hline
\textbf{Models} & \textbf{Regression (MLP) Head} & \textbf{FC Layer + MLP} & \textbf{Complete Model} \\
\hline
GCN & 93.725 (1.716) & 82.737 (1.119) & 75.621 (3.456) \\
GAT & 91.323 (5.436) & 83.671 (3.974) & 72.195 (1.112) \\
GIN & 60.960 (0.158) & 55.011 (0.474) & 52.914 (1.218) \\
GraphSAGE & 84.543 (1.924) & 75.237 (0.264) & 70.752 (2.750) \\
\hline
\end{tabular}
\end{table}

\subsection{Explaining GNN Model Prediction Using GNNExplainer}
\begin{figure}[H]
\centering
\includegraphics[width=0.95\textwidth]{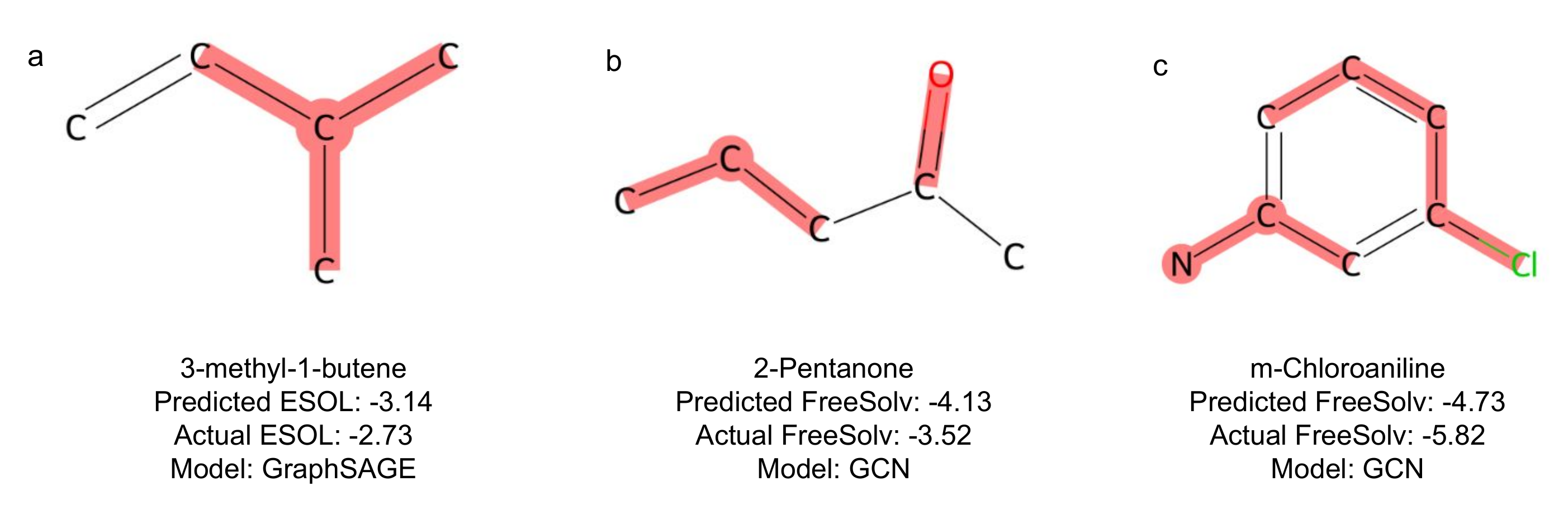}
\caption{\textbf{Explanation of molecular graphs.} (a-c) Graph explanation for three case-study molecules with actual and predicted target values.}
\label{fig:fig5}
\end{figure}

ML models are generally trained using fingerprint (ECFP4) embeddings; these embeddings are difficult to draw bits (fingerprints) level explanations about model’s prediction using existing libraries like SHAP \cite{Rodrguez-Prez2020}. GNNs are trained on molecular graphs and they also support libraries like GNNExplainer, to directly draw graph level explanations (Figure \ref{fig:fig5}).  In Figure \ref{fig:fig5}, we have plotted the explanation of 3 (selected for simplicity and smaller size) case study molecules. As can be seen in Figure. \ref{fig:fig5}a the highlighted region belongs to the hydrophobic core, which negatively contributes to aqueous solubility (ESOL) of 3-methyl-1-butene. In Figure \ref{fig:fig5}b the two highlighted regions in the 2-pentanone molecule, first the carbonyl group that increases aqueous solubility, whereas the hydrophobic tail reduces the water solubility. In case of m-chloroaniline (Figure \ref{fig:fig5}c), the amine group containing nitrogen (N) atom positively contributes, whereas the aromatic ring and chlorine (Cl) atom negatively contributes to its aqueous solubility. We initially wanted to compare the task-specific molecular features, but no molecule (from 7 case studies) has absolute error below SOTA on both the target values (e.g. ESOL and FreeSolv). Therefore, we only draw explanations for those, with absolute error below SOTA. This was done to infer high confidence explanations.

We found that fingerprints are non-complementary in nature, with graph embeddings; since no gain in performance was seen except for FreeSolv data. This is concordant with previous findings by Ha et al., 2025, where authors have demonstrated that concatenating fingerprints to a superior GNN model can reduce the performance \cite{Ha2025}. In case of FreeSolv, GNN embeddings alone couldn’t capture complete representation, probably due to smaller dataset thereby concatenating fingerprints boosted the performance. In general, GNN are highly scalable architectures, in comparison to their conventional ML/DL counterparts, with higher performance to be expected as dataset volume expands \cite{Sypetkowski2024}. Additionally, the top performing GNN models were GCN or GraphSAGE, whereas based on training efficiency, GCN and GIN were the fastest trainers.  GIN was frequently the second-best model in terms of absolute performance in RMSE (Table \ref{tab:tab2}). GIN was also the top performer in transfer learning tasks. These results suggest that if training time or computational resources is a constraint, then GCN or GIN can be a better tradeoff between top vs optimal performance. Additionally, we have observed through error analysis that prediction errors of GNN are systematic, with high error molecules mostly with larger size and more rings.

\section{Conclusions}
\label{sec:Conclusions}
In conclusion, our study demonstrates that GNN (GraphSAGE and GCN) models trained on molecular graphs can give superior performance over ML models trained using fixed-length molecular fingerprints (ECFP4) for regression tasks. Additionally, the quality of prediction made by GNN models was also better i.e. lower percentage of molecules with high absolute error. Since GNN models are trained using molecular graphs, they can be used for transfer learning objectives to achieve significantly higher performance as shown in our study. The study also highlights the fact that choice of the GNN layers can also provide significant boost in performance and, if compute is a constraint, then GCN or GIN can be a better tradeoff.  There are some limitations as well, unlike ML models where fixed-size fingerprints can be generated, GNN models showed lower predictive performance for large and complex molecules. The training time for GNN models is significantly higher as compared to ML models, which can possess scalability issues in resource constrained environments. Nevertheless, with better and more efficient architecture these limitations can be overcome. 

%Bibliography
\bibliographystyle{unsrt}  
\bibliography{references}

\end{document}